# Learning Subject-Specific Anatomical Representations via Manifold Expansion: Application to Accelerated Multi-Contrast MRI

Ruimin Feng[a,b], Wanyu Bian[a,b], Albert Jang[a,b], Zachary Stewart[b,c], and Fang Liu[a,b]

[a] Athinoula A. Martinos Center for Biomedical Imaging, Massachusetts General Hospital, Charlestown, Massachusetts, United States

[b] Harvard Medical School, Boston, Massachusetts, United States

[c] Department of Radiology, Massachusetts General Hospital, Charlestown, Massachusetts, United States

| | |
|---|---|
| Running title: | Anatomical Representation Learning via Manifold Expansion |
| Corresponding author: | Fang Liu, Ph.D.<br>Athinoula A. Martinos Center for Biomedical Imaging<br>149 13th Street, Suite 2301<br>Charlestown, Massachusetts 02129<br>United States |
| Phone: | 1-617-643-4887 |
| Fax: | 1-617-726-7422 |
| Email: | fliu12@mgh.harvard.edu |
| Word count: | 5664 |
| Key words (3 to 6): | anatomical representation learning, manifold expansion, disentangled representation learning, multi-contrast MRI, implicit neural representation |
| Date: | October 2, 2026 |

## ABSTRACT

**Purpose:** Clinical MRI routinely acquires multiple contrast-weighted images of the same anatomy for complementary tissue characterization. However, current accelerated MRI methods typically reconstruct each contrast independently, without fully exploiting shared anatomical information. This work aims to learn anatomical representations invariant to contrast-dependent appearance for reconstruction of accelerated multi-contrast MRI.

**Methods:** We propose MAX (MAnifold eXpansion), a subject-specific framework that learns anatomical representations from a single fully sampled reference contrast. To address the under-constrained separation of shared anatomy and contrast-dependent components from a single image, MAX expands the multi-contrast manifold using anatomy-preserving intensity augmentations. A disentangled implicit neural representation models augmented samples using shared spatial coordinates for anatomy and spatially invariant coordinates for contrast appearance. The learned anatomical representation is then fixed, with the contrast representation adapted to the undersampled target data, followed by unrolled refinement. Theoretical analyses further provide insight into the disentangled representation learning and explain how the learned anatomical representation improves the target contrast reconstruction.

**Results:** At R = 8 for brain MRI and R = 6 for knee MRI, MAX achieves the highest mean PSNR and SSIM across all tasks, improving PSNR by more than 1 dB over the strongest baseline for both brain contrasts. MAX more faithfully recovers subtle anatomical and pathological structures and remains robust to inter-contrast motion, structural heterogeneity between reference and target contrasts, and measurement noise.

**Conclusion:** MAX provides a general strategy for leveraging high-quality reference scans in accelerated MRI and has the potential to be extended to other reference-assisted MRI inverse problems.

## 1. Introduction

Clinical MRI routinely acquires multiple contrast-weighted images to reveal complementary tissue properties.[1] Although generated by distinct contrast mechanisms, these images represent different observations of a common anatomical substrate. This shared anatomy can provide a powerful source of prior information for a broad range of cross-contrast imaging tasks, such as fast MRI, distortion correction, denoising, and super-resolution. However, anatomical information is intertwined with contrast-specific appearance in the observed images, making direct transfer susceptible to introducing irrelevant features or suppressing patterns unique to the target contrast. A fundamental challenge in multi-contrast MRI is therefore to selectively exploit shared anatomical information without compromising contrast-specific features.

For reconstructing accelerated MRI, existing methods have sought to utilize complementary information across MRI contrasts. Conventional methods encode inter-contrast consistency using predefined mathematical assumptions,[2–7] including hierarchical Bayesian models,[2] joint total variation and group sparsity,[3] directional gradient guidance,[5] and coupled dictionary learning.[7] Although these methods provide explicit control over what information is shared, their prescribed assumptions capture only limited forms of anatomical correspondence and remain susceptible to contrast-dependent variations between tissues. More recently, deep learning methods have learned richer cross-contrast features from multi-contrast datasets.[8–20] Representative examples include deep unregistered networks that jointly learn image registration and reconstruction from reference contrasts,[13] transformer-based architectures for dual-domain feature fusion,[14,15] decomposition-based variational networks with designed shared and contrast-specific branches,[17] and multi-task meta-learning frameworks that learn mutual features across MRI contrasts.[20] However, these learned features are derived from large-scale training datasets and remain largely implicit in network parameters, making it difficult to model and selectively exploit the shared anatomical information. Moreover, cross-contrast relationships learned from training datasets may not generalize well with variations in anatomy, acquisition protocols, or measurement noise. Therefore, existing approaches face a fundamental trade-off: handcrafted priors are controllable but limited in expressiveness, whereas learned priors are expressive but difficult to interpret and control. Overcoming this trade-off requires a defined anatomical representation that can be learned for the individual examination and separated from contrast-specific appearance.

To address these limitations, we propose MAX (MAnifold eXpansion), a subject-specific framework that learns shared anatomical representations from a single fully sampled reference-contrast image to reconstruct highly accelerated target-contrast MRI. Because a single image provides only one observation of the underlying anatomy, disentangling shared anatomy from contrast-dependent appearance is inherently under-constrained. MAX overcomes this challenge by expanding the local multi-contrast manifold through anatomy-preserving intensity augmentations, generating diverse observations that facilitate cross-contrast representation learning while preserving the underlying spatial structure. To explicitly disentangle anatomy from image appearance, MAX employs the disentangled representation learning strategy,[21–23] implemented using an implicit neural representation (INR).[24,25] Specifically, the INR represents shared anatomy as a function of spatial coordinates while modeling contrast-dependent appearance with a low-dimensional, spatially invariant latent embedding. The learned anatomical representation is then incorporated into a physics-guided reconstruction framework through representation-constrained adaptation, followed by unrolled refinement to recover target-specific structures. Because the framework is subject-specific, it learns personalized anatomical representations directly from the reference scan, eliminating the need for external training datasets and reducing the risk of distribution shift. We further provide a theoretical interpretation of the proposed framework and evaluate it on multiple reference-target contrast pairs in both brain and knee MRI. Experimental results demonstrate consistent improvements in reconstruction accuracy and robustness over existing state-of-the-art methods.

## 2. Theory

**Figure 1(A)** provides a conceptual illustration of the proposed MAX framework. In this section, we present its theoretical foundation based on a manifold model. This analysis motivates the key components of MAX, including disentangled representation learning, local multi-contrast manifold expansion from a single fully sampled reference contrast, and physics-guided adaptation and refinement for target-contrast reconstruction.

### 2.1 Problem Formulation

Given a fully sampled reference-contrast image $\mathbf{x}_{\text{ref}}$ and an undersampled target-contrast acquisition of the same subject and anatomical region, let $\mathbf{x}_{\text{tar}} \in \mathbb{C}^D$ denote the target image, where $D$ denotes the total number of image voxels. The acquired target k-space measurements are

modeled as:

$$\mathbf{y}_{\text{tar}} = \mathbf{A}\mathbf{x}_{\text{tar}} + \boldsymbol{\varepsilon}, \quad (1)$$

where $\mathbf{A}$ represents the MRI encoding operator incorporating coil sensitivity maps, Fourier transformation, and k-space undersampling, and $\boldsymbol{\varepsilon}$ denotes measurement noise. The objective is to reconstruct $\mathbf{x}_{\text{tar}}$ from $\mathbf{y}_{\text{tar}}$ by exploiting the shared anatomical information contained in $\mathbf{x}_{\text{ref}}$.

## 2.2 Manifold Model for Multi-Contrast MRI

Although multi-contrast MRI images are acquired with different contrast mechanisms and exhibit distinct image appearances, they depict the same underlying anatomy. We therefore assume that these images lie approximately on a low-dimensional manifold embedded in the high-dimensional ambient image space.[26–28] Let $\mathcal{M} \subset \mathbb{R}^D$ denote the subject-specific multi-contrast image manifold. Accordingly, for the $k$-th contrast-weighted image, $\mathbf{x}_k$ , can be regarded as a point on the manifold:

$$\mathbf{x}_k \in \mathcal{M}, \dim(\mathcal{M}) = d \ll D. \quad (2)$$

By definition, a $d$-dimensional manifold is locally equivalent to $\mathbb{R}^d$.[29] Specifically, there exists a neighborhood $U \subset \mathcal{M}$ and a local parameter domain $V \subset \mathbb{R}^d$, together with a local coordinate chart (parameterization) $\mathcal{F}: V \to U$, such that:

$$\mathbf{x}_k = \mathcal{F}(\boldsymbol{\xi}_k), \boldsymbol{\xi}_k \in V. \quad (3)$$

Here, $\boldsymbol{\xi}_k$ denotes the latent variable associated with the $k$-th contrast. Furthermore, motivated by the observation that different contrasts share the same underlying anatomy while exhibiting contrast-dependent appearance variations, we decompose the latent variable into shared and contrast-specific components:

$$\boldsymbol{\xi}_k = (\boldsymbol{\eta}, \boldsymbol{\gamma}_k), \quad (4)$$

where $\boldsymbol{\eta}$ determines the anatomical information shared across contrasts and $\boldsymbol{\gamma}_k$ captures the contrast-specific image appearance, including tissue-dependent signal weighting and intensity distribution. The image formation model is therefore written as:

$$\mathbf{x}_k = \mathcal{F}(\boldsymbol{\eta}, \boldsymbol{\gamma}_k). \quad (5)$$

We now consider the practical setting in which a fully sampled reference contrast and an undersampled target acquisition are available. We assume that they reside within the same neighborhood of the subject-specific multi-contrast manifold and therefore share a common local parameterization $\mathcal{F}$. Under this assumption, if the shared anatomical component $\boldsymbol{\eta}$ can be reliably estimated from the reference image, target reconstruction reduces to estimating only the contrast-specific latent variable $\boldsymbol{\gamma}_{\text{tar}}$:

$$\hat{\boldsymbol{\gamma}}_{\text{tar}} = \underset{\boldsymbol{\gamma}_{\text{tar}}}{\operatorname{argmin}} \parallel \mathbf{A}\left[\mathcal{F}(\boldsymbol{\eta}, \boldsymbol{\gamma}_{\text{tar}}) \odot e^{i\boldsymbol{\phi}_{\text{tar}}}\right] - \mathbf{y}_{\text{tar}} \parallel_2^2. \tag{6}$$

where $\boldsymbol{\phi}_{\text{tar}}$ denotes the target image phase, which is modeled separately because the manifold formulation above considers only the magnitude image. Compared with directly reconstructing the full image, Eq. (6) reduces the unknowns from the high-dimensional image space to the lower-dimensional contrast-specific latent variable $\boldsymbol{\gamma}_{\text{tar}}$, thereby improving reconstruction stability under aggressive undersampling. The remaining challenges are (1) learning an effective approximation of the manifold mapping $\mathcal{F}$, and (2) accurately estimating the shared anatomical representation $\boldsymbol{\eta}$ from only a single fully sampled reference image. The proposed method addresses these challenges through the manifold learning framework described below.

### 2.3 Realizing Manifold Mapping with a Disentangled Implicit Neural Representation

The manifold formulation in Section 2.2 establishes the existence of a local parameterization $\mathcal{F}$ that maps latent variables to the image space. Since the analytical form of $\mathcal{F}$ is generally unknown, we approximate it using a disentangled implicit neural representation, which provides a flexible nonlinear function approximator for continuous image representations. Specifically, the shared anatomical component $\boldsymbol{\eta}$ is represented as a function of spatial coordinates to encode anatomical information common across all contrasts, whereas the contrast-specific component $\boldsymbol{\gamma}_k$ is represented by a spatially invariant contrast coordinate that modulates image appearance without altering the underlying anatomy:

$$\boldsymbol{\eta} = h_{\psi_s}(p_i), \boldsymbol{\gamma}_k = h_{\psi_c}(c_k), \tag{7}$$

where $p_i \in \Omega \subset \mathbb{R}^3$ denotes the spatial coordinate of the $i$-th voxel, with $i = 1,2, \dots, D$; $c_k \in \mathbb{R}^{d_c}$ denotes a contrast coordinate associated with the $k$-th contrast, and $h_{\psi_s}$ and $h_{\psi_c}$ are the corresponding encoders with parameters $\psi_s$ and $\psi_c$. The resulting embeddings are concatenated along the feature dimension and passed to a multilayer perceptron (MLP), $f_\theta$, which defines the nonlinear manifold mapping. The resulting manifold parameterization is therefore expressed as:

$$\mathcal{F}(\boldsymbol{\eta}, \boldsymbol{\gamma}_k) = f_\theta\left(\text{concat}\left(h_{\psi_s}(p_i), h_{\psi_c}(c_k)\right)\right). \tag{8}$$

To provide further theoretical insight, the **Supporting Information ("Local Linear Interpretation of the Learned Manifold Parameterization")** shows that, under a first-order local approximation, the learned disentangled INR admits the following image representation:

$$\mathbf{x}_k \approx \sum_{r=0}^{r_c} \mathbf{b}_r \, \mathrm{a}_r^{(k)}, \tag{9}$$

where $\mathbf{b}_r$ denote locally defined basis images induced by the local linearization of the learned manifold mapping, $\mathrm{a}_r^{(k)}$ are contrast-dependent low-dimensional coefficients, and $r_c$ denotes the dimension of the contrast embedding. This result demonstrates that the learned nonlinear manifold parameterization is locally equivalent to a low-dimensional basis-coefficient representation, thereby establishing a theoretical connection between the proposed framework and conventional low-rank or subspace MRI reconstruction models.[30–32]

## 2.4 Learning Anatomical Representations via Reference-Based Manifold Expansion

With only a single fully sampled reference contrast, learning the shared anatomical representation and the contrast-specific representation is an underdetermined problem. Specifically, multiple combinations of the spatial encoder $h_{\psi_s}$, contrast encoder $h_{\psi_c}$, and mapping network $f_\theta$ can reproduce the same reference image while yielding different decompositions of anatomical and contrast-specific information. To reduce this ambiguity, we expand the local multi-contrast manifold by generating additional observations from the fully sampled reference image. These augmented samples preserve the underlying anatomy while introducing appearance variations that simulate those observed across MRI contrasts:

$$\mathcal{X}_{\text{aug}} = \{\mathbf{x}_\alpha\}_{\alpha \in \Lambda_{\text{aug}}}. \tag{10}$$

The augmented observations provide additional constraints for learning a disentangled representation. Accordingly, the parameters of the spatial encoder, contrast encoder, and mapping network are optimized using all augmented observations $\mathcal{X}_{\text{aug}}$:

$$(\psi_s^\star, \psi_c^\star, \theta^\star) = \arg\min_{\psi_s, \psi_c, \theta} \sum_{\alpha \in \Lambda_{\text{aug}}} \sum_{i=1}^{D} \left\| f_\theta\left(\text{concat}\left(h_{\psi_s}(p_i), h_{\psi_c}(c_\alpha)\right)\right) - \mathbf{x}_\alpha(p_i) \right\|_2^2. \tag{11}$$

Jointly fitting $\mathcal{X}_{\text{aug}}$ encourages the spatial encoder to learn anatomy that is invariant across appearance changes, while the contrast encoder captures only appearance-specific variations.

The effectiveness of this framework depends critically on the choice of augmentation strategy. Under the local linear approximation established in Eq. (9), effective augmentations should preserve the shared anatomy while generating appearance variations that span the local neighborhood of the target contrast. To quantify this property for theoretical analysis and offline augmentation selection, we define the following augmentation coverage error. Let $\mathbf{B}_{\text{aug}} \in$ $\mathbb{R}^{D\times|\Lambda_{\mathrm{aug}}|}$ denote the matrix whose columns are the vectorized augmented images:

$$e_{\mathrm{aug}} \coloneqq \min_{\boldsymbol{\omega}} \| \mathbf{x}_{\mathrm{tar}} - \mathbf{B}_{\mathrm{aug}}\boldsymbol{\omega}\|_2. \tag{12}$$

Here, $\mathbf{x}_{\mathrm{tar}}$ denotes a fully sampled target image available only during offline augmentation selection (see **Section 3.2**) and $\boldsymbol{\omega}$ denotes the coefficients used to linearly combine the augmented observations. A smaller $e_{\mathrm{aug}}$ indicates that the augmented observations provide a more complete local spanning set for the appearance variations associated with the target contrast, thereby facilitating disentangled representation learning. Because Eq. (12) is derived from the local linear approximation of the nonlinear manifold, it serves as a practical criterion for analyzing and comparing augmentation strategies rather than an exact characterization of the nonlinear representation error.

## 2.5 Physics-Guided Target Contrast Reconstruction with Learned Anatomical Representations

Once the anatomical representation has been learned, the spatial encoder $h_{\psi_s^*}$ can be fixed, while the contrast encoder $h_{\psi_c}$ and MLP $f_\theta$ are adapted to the target image by optimizing:

$$(\psi_c^{\mathrm{tar}}, \theta^{\mathrm{tar}}) = \underset{\psi_c,\theta}{\operatorname{argmin}} \left\| \mathbf{A} f_\theta \left( \mathrm{concat}\left( h_{\psi_s^*}(\mathbf{p}), h_{\psi_c}(c_{\mathrm{tar}}) \right) \right) - \mathbf{y}_{\mathbf{tar}} \right\|_2^2, \tag{13}$$

where $\mathbf{p} = [p_1, p_2, \ldots, p_D]^T \in \mathbb{R}^{D\times 3}$ denotes the collection of all spatial coordinates, and $c_{\mathrm{tar}}$ is the contrast coordinate corresponding to the target image. Fixing the spatial encoder reduces the trainable degrees of freedom, thereby improving reconstruction stability under aggressive undersampling.

Although the adapted INR provides a subject-specific reconstruction consistent with the learned manifold, its representational capacity may be insufficient to capture subtle target-specific structures outside the manifold. Therefore, the INR reconstruction, $\mathbf{x}_{\mathrm{inr}}$, is further refined using a physics-guided unrolled reconstruction framework[33] initialized with $\mathbf{x}_0 = \mathbf{x}_{\mathrm{inr}}$. Each iteration alternates between learned CNN refinement:

$$\mathbf{z}^{(n)} = \mathcal{D}_\psi\left(\mathbf{x}^{(n)}\right), \tag{14}$$

and a physics-guided data-consistency update with INR-anchored regularization:

$$\mathbf{x}^{(n+1)} = \underset{\mathbf{x}}{\operatorname{argmin}} \| \mathbf{A}\mathbf{x} - \mathbf{y}_{\mathrm{tar}}\|_2^2 + \lambda \| \mathbf{x} - \mathbf{z}^{(n)}\|_2^2 + \mu \| \mathbf{x} - \mathbf{x}_{\mathrm{inr}}\|_2^2. \tag{15}$$

Here, $\lambda$ and $\mu$ control the CNN-based and INR-anchored regularization, respectively. This refinement retains reliable INR-derived anatomical information while recovering target measurement-supported features absent from the learned manifold. **The Supporting Information ("Error Analysis of Physics-Guided Unrolled Refinement")** further analyzes the proposed refinement scheme and shows that an accurate INR initial estimate combined with the INR-anchor regularization reduces the upper bound on the final reconstruction error.

## 3. Methods

### 3.1 Overview of the Proposed Method

**Figure 1(B)** shows the detailed architecture of MAX, which comprises two stages: (1) anatomical representation learning and (2) target contrast reconstruction. The encoders $h_{\psi_s}$ and $h_{\psi_c}$ are implemented using hash encoding,[34] which maps input coordinates to learnable multi-resolution grids. This improves the capacity of the INR to represent high-frequency information with compact computational cost.

#### 3.1.1 Anatomical Representation Learning on Augmented Data

In this stage, the spatial hash encoder, contrast hash encoder, and MLP are jointly trained on the augmented dataset. All augmented images share the same spatial coordinates $\{p_i\}_{i=1}^{D}$, which are defined on a regular Cartesian grid, while each image is assigned a randomly sampled contrast coordinate. The network is optimized to fit the augmented images using the following loss function:

$$\mathcal{L}_{\text{prior}} = \frac{1}{N_{\text{aug}}} \sum_{\alpha=1}^{N_{\text{aug}}} \| \hat{\mathbf{x}}_\alpha - \mathbf{x}_\alpha \|_2^2, \tag{16}$$

where $N_{\text{aug}}$ is the number of augmented images, $\hat{\mathbf{x}}_\alpha$ denotes the INR prediction for the $\alpha$-th augmented sample over all voxel coordinates, and $\mathbf{x}_\alpha$ is the corresponding ground truth augmented image.

#### 3.1.2 Reconstruction of Target Contrast

In the reconstruction stage, the learned spatial hash encoder from Stage 1 is fixed, whereas the contrast encoder and MLP are adapted to the undersampled target measurements. Additionally, the target phase is modeled as a smooth implicit function by a separate low-capacity branch. This branch combines low-frequency positional encoding[24] with a lightweight MLP to map spatial coordinates to a unit-magnitude complex phase term. The learnable components are optimized by enforcing data consistency with the acquired target k-space measurements:

$$\mathcal{L}_{\text{inr}} = \|\mathbf{A}\mathbf{x}_{\text{inr}} - \mathbf{y}_{\text{tar}}\|_2^2. \tag{17}$$

The resulting $\mathbf{x}_{\text{inr}}$ serves as the initial estimate and the anchor for the unrolled refinement.

The unrolled module implements Eq. (14) using a CNN composed of residual blocks[35] and solves the update in Eq. (15) using conjugate gradient. Following ZS-SSDU,[36] the acquired k-space samples are split into Θ, Λ, and Γ for data consistency, self-supervised training, and validation, respectively. The refinement network is optimized by the following self-supervised loss evaluated on Λ:

$$\mathcal{L}_{\text{refine}} = \frac{1}{2}\frac{\left\|\mathbf{A}_\Lambda \mathbf{x}^{(N)} - \mathbf{y}_\Lambda\right\|_2}{\|\mathbf{y}_\Lambda\|_2 + \epsilon} + \frac{1}{2}\frac{\left\|\mathbf{A}_\Lambda \mathbf{x}^{(N)} - \mathbf{y}_\Lambda\right\|_1}{\|\mathbf{y}_\Lambda\|_1 + \epsilon}, \tag{18}$$

where $\mathbf{A}_\Lambda$ and $\mathbf{y}_\Lambda$ denote the forward model and measured k-space data restricted to Λ, respectively; $N$ is the number of unrolled iterations, and $\epsilon$ is a small constant to avoid division by zero.

### 3.2 Detailed Strategy for Reference-Based Data Augmentation and Selection

Following the augmentation framework in Section 2.4, anatomy-preserving intensity transformations are used to expand the multi-contrast manifold neighborhood around the reference image. First, a large candidate bank is generated from the fully sampled reference image using global affine intensity changes, contrast inversion, and nonlinear gamma, logarithmic, sigmoid, piecewise-linear, and polynomial mappings. Transformation parameters are randomly sampled from predefined ranges (see **Table S1**). In addition to the base transformations, the bank also includes higher-order augmentations formed by randomly composing two to six operations from the same transformation pool, providing both simple and complex contrast variations. Second, a compact transformation subset is selected greedily according to Eq. (12). For each reference-target contrast pair, selection is performed offline using a representative fully sampled image pair; at each iteration, the transformation yielding the largest reduction in $e_{\text{aug}}$ is added. The selected subset is then fixed for all subsequent experiments involving the same contrast pair.

### 3.3 Motion Correction for Inter-Contrast Misalignment

Multi-contrast reconstruction is susceptible to inter-contrast spatial misalignment caused by subject motion. To address this issue, we introduce a motion-correction (MAX-MoCo) strategy. First, a coarse target image is reconstructed from the central region of the target k-space and serves as the fixed image for registration of the reference contrast. The resulting transformation provides the initial estimate of a learnable motion model. During INR reconstruction, the INR output is spatially transformed by this motion model, while its motion parameters are jointly optimized together with the INR parameters by minimizing $\mathcal{L}_{\text{inr}}$. This strategy allows MAX to compensate for spatial misalignment between the reference and target contrasts. In this study, we employed a six-degree-of-freedom rigid motion model for the corresponding experiment.

### 3.4 Implementation Details

The proposed framework was implemented in Python 3.9 using PyTorch 2.8.0. The hash encoding was realized based on the tiny-cuda-nn library (https://github.com/NVlabs/tiny-cuda-nn), with all encoding parameters set to default values. The phase branch employed positional encoding with four frequency bands. Both MLPs consisted of three hidden layers with 64 neurons per layer and ReLU activations. No activation was applied to the output layer. The INR was optimized for 1000 iterations using the Adam optimizer[37] with a learning rate of $1 \times 10^{-3}$.

The unrolled refinement network comprised an input convolution, 15 residual blocks with 64 channels and ReLU activations, and an output convolution predicting residual corrections. It was unfolded into eight unrolled blocks and optimized using Adam with a learning rate of $5 \times 10^{-4}$. Early stopping was applied after 25 epochs without improvement in validation loss. The acquired k-space samples were randomly partitioned into $\Theta$, $\Lambda$, and $\Gamma$ with ratios of 0.55, 0.40, and 0.05, respectively. A small validation set ($\Gamma$) was used to preserve more measurements for training and data consistency. The weighting parameter $\lambda$ in Eq. (15) was learned during optimization. Except under low-SNR conditions, the INR-anchored regularization parameter $\mu$ was set to 0, as a sufficiently faithful INR initialization was found to be more beneficial than explicitly constraining the subsequent optimization. In the SNR robustness experiment (**Section 3.6.2**), $\mu$ was set to 0.02 and 0.05 at 20 and 10 dB, respectively, to strengthen INR guidance based on the high-SNR reference and reduce the influence of target-measurement noise. For manifold expansion, 1000 augmented images formed the candidate bank, from which 30 were selected by greedy search.

All experiments were conducted on a workstation equipped with an Intel(R) Xeon(R) Platinum 8562Y+ CPU and an NVIDIA H100 GPU with 96 GB of memory. The source code will be released at https://github.com/I3Tlab/MAX-Reconstruction upon publication.

### 3.5 Image Datasets

We conducted experiments on two anatomical datasets: brain and knee. Both datasets were collected with approval from the institutional ethics committee, and written informed consent was obtained from all subjects before data acquisition. The brain dataset included 10 healthy subjects with fully sampled k-space data acquired for MPRAGE, T2-weighted, and FLAIR. The corresponding brain images had an in-plane resolution of $1 \times 1$ mm$^2$ and a slice thickness of 3 mm. The knee dataset included 18 patients with knee osteoarthritis, with fully sampled k-space data acquired for T2-weighted fat-suppressed and T1-weighted non-fat-suppressed contrasts. The corresponding knee images had an in-plane resolution of 0.4×0.4 $mm^2$ and a slice thickness of 4.4 mm.

### 3.6 Experimental Setup

We conducted a series of experiments to comprehensively assess the proposed MAX method.

#### 3.6.1 Comparison with State-of-the-Art Methods

These experiments evaluated the reconstruction performance of MAX against representative state-of-the-art methods. The comparison methods included both single-contrast reconstruction approaches and reference-guided multi-contrast reconstructions. The single-contrast baselines are compressed sensing-based L1-ESPIRiT,[38] the INR-based IMJENSE,[25] and the zero-shot unrolled method ZS-SSDU.[36] The reference-guided methods are dTVMRI,[5] which exploits structural similarity across contrasts through joint total variation regularization with directional information, and CDLMRI,[7] which uses dictionary learning to capture shared and contrast-specific features. Because dTVMRI and CDLMRI are originally formulated for single-channel magnitude images rather than complex-valued multi-coil acquisitions, the fully sampled multi-coil data were coil-combined into magnitude images and retrospectively undersampled in k-space to generate their reconstruction inputs. All comparison methods were implemented using the authors' publicly available source code, with hyperparameters fine-tuned on our datasets for optimal performance.

For the brain dataset, MPRAGE was used as the reference contrast, while T2-weighted and FLAIR images were reconstructed as target contrasts. The target k-space data were retrospectively undersampled along one phase-encoding direction with an acceleration rate of R = 8. The knee dataset was used to assess the recovery of pathological features, with non-fat-suppressed T1-weighted images as the reference and fat-suppressed T2-weighted images as the target. The knee k-space data were retrospectively undersampled at R = 6. Reconstruction performance was evaluated using peak signal-to-noise ratio (PSNR) and structural similarity index measure (SSIM). A reader study was conducted on the knee dataset in which an experienced musculoskeletal radiologist specialized in knee MRI evaluated the reconstructions from all the methods. Images were presented randomly and scored on a 5-point Likert scale[39] (1=very poor, 5=excellent; higher scores indicate better quality) across four criteria: artifact suppression, perceived sharpness, diagnostic confidence, and overall image quality.

### 3.6.2 Robustness Evaluation

These experiments evaluated the robustness of MAX under practical perturbations, including inter-contrast motion, intrinsic structural heterogeneity between contrasts, and measurement noise in the target k-space data.

(1) Inter-contrast motion: After the regular brain scan, five subjects were instructed to move their heads before acquiring an additional MPRAGE. The displaced MPRAGE image was used as the reference contrast, while the original-position FLAIR acquisition served as the target contrast. To evaluate the proposed motion-correction strategy, the initial INR reconstruction (MAX-Initial) and the final reconstructed images obtained with and without motion correction were compared, and the corresponding reference-target spatial alignment was visualized. Furthermore, quantitative reconstruction performance was evaluated across all five subjects.

(2) Intrinsic structural heterogeneity: Synthetic lesion-like abnormalities were introduced into the target FLAIR image while the reference remained unchanged. Specifically, the original multi-coil k-space data were first coil-combined into a complex image using the estimated sensitivity maps. A predefined lesion mask $\mathbf{M}$ was then applied to locally increase the image intensity according to $\mathbf{x}_{\text{lesion}} = \mathbf{x}_{\text{ori}}(1 + \rho\mathbf{M})$, where $\rho = 0.6$. The modified image was projected back to multi-coil k-space using the sensitivity maps and Fourier transform, followed by the addition of the original k-space residual to preserve measurement noise. The resulting data were retrospectively undersampled at R = 8 and reconstructed using dTVMRI, CDLMRI, MAX-Initial, and MAX for comparison.

(3) Measurement noise: Complex Gaussian noise was added to the undersampled target k-space to generate measurements at SNRs of 30, 20, and 10 dB, while the reference contrast remained unchanged. Reconstructions at each SNR were compared with the corresponding fully sampled images to evaluate the robustness of MAX to measurement noise.

### 3.6.3 Ablation Studies

These experiments evaluated the contribution of key components in MAX.

(1) Data augmentation for representation learning. To evaluate the effectiveness of the proposed data augmentation strategy, we compared two variants of the representation learning framework at Stage 1: one trained using a single reference image (MAX-Single) and the other trained using the augmented dataset (MAX-Augmented). Specifically, the contrast embeddings learned by MAX-Single and MAX-Augmented were visualized using t-SNE[40] to examine the organization of the embedding space. Three additional contrast embeddings were randomly sampled from the learned embedding space, concatenated with the anatomical embeddings, and passed through the MLP decoder to evaluate whether the learned manifold could generate diverse contrast appearances. In addition, the target-contrast images reconstructed directly from the learned representations were compared qualitatively and quantitatively to evaluate the impact of manifold expansion on target-contrast adaptation.

(2) INR-based initial estimate. To assess how the initial estimate affects the final reconstruction, we compared three starting points for unrolled refinement at an extreme acceleration factor of R = 12: the zero-filled image, the reference image, and the proposed INR-based initialization. The corresponding initial estimates and final reconstructions were compared qualitatively.

# 4. Results

## 4.1 Reconstruction Results of Different Methods

**Figure 2** shows representative T2-weighted reconstruction results for two subjects at R = 8, using MPRAGE as the reference contrast. As highlighted in the enlarged views, the zero-filled images exhibit severe artifacts. Although L1-ESPIRiT and IMJENSE suppress aliasing, both produce blurred tissue boundaries and lose high-frequency details. By exploiting the reference contrast, dTVMRI better preserves major anatomical structures but introduces patchy textures and piecewise-constant patterns. CDLMRI recovers more realistic structures but retains noticeable artifacts. ZS-SSDU further improves structural recovery through learned CNN priors, although residual aliasing and over-smoothing remain. In contrast, MAX produces reconstructions closest to the fully sampled images, recovering subtle anatomical details and sharp tissue interfaces in both deep gray matter and cortical regions. The error maps confirm these observations, with MAX consistently showing the lowest reconstruction errors.

**Figure 3** presents representative FLAIR reconstruction results at R = 8. Similar trends to those in **Figure 2** are observed. Single-contrast methods (L1-ESPIRiT, IMJENSE, and ZS-SSDU) exhibit blurred boundaries and residual artifacts, whereas reference-guided multi-contrast methods (dTVMRI and CDLMRI) improve structural recovery but remain affected by patchy textures or artifacts. In comparison, MAX more faithfully reconstructs fine tissue structures, with the error maps indicating the smallest deviations from the fully sampled target.

**Figure 4** shows representative knee reconstructions for two subjects at R = 6, using a fully sampled non-fat-suppressed T1-weighted image as the reference. The enlarged views highlight pathology-related structures. In Example 1, the yellow arrows indicate subchondral cysts of the tibia. The cysts are poorly delineated in the zero-filled reconstruction and remain less well-defined with L1-ESPIRiT, IMJENSE, dTVMRI, and CDLMRI. ZS-SSDU and MAX provide a better depiction of the cysts, with MAX exhibiting the sharpest cyst boundaries. The orange arrows indicate superficial cartilage fissuring, for which MAX most faithfully preserves the lesion morphology. MAX also better preserves the surrounding anatomical boundaries (blue arrows). In Example 2, the green arrows indicate a macerated and extruded medial meniscus. The lesion boundaries remain degraded in the zero-filled, L1-ESPIRiT, IMJENSE, and dTVMRI results. CDLMRI and ZS-SSDU improve visualization but still exhibit artifacts, whereas MAX most accurately recovers the lesion extent and surrounding anatomy. The error maps further show that MAX consistently produces the smallest errors around the pathological regions.

**Figure 5** summarizes the statistical analysis of reconstruction performance on the brain and knee datasets. In **Figures 5(A) and 5(B)**, MAX yields the highest median PSNR and SSIM for both T2-weighted and FLAIR reconstruction, with the smallest inter-subject variation. For the knee dataset (**Figure 5(C)**), MAX provides the highest median PSNR, while its median SSIM is comparable to that of ZS-SSDU. In the reader study (**Figure 5(D)**), MAX receives consistently high ratings across all four criteria, including the highest score for artifact suppression and perceived sharpness. **Table S2** reports the corresponding mean and standard deviation of PSNR and SSIM. Compared with the second-best method, MAX improves mean PSNR by 1.56, 1.39, and 0.17 dB and mean SSIM by 0.010, 0.008, and 0.001 for brain T2-weighted, brain FLAIR, and knee reconstruction, respectively.

### 4.2 Results of Robustness Evaluation

**Figure 6** presents reconstructions at R = 8 with and without the proposed motion correction strategy under motion-induced misalignment between the reference and target contrasts. Without MoCo (first row, **Figure 6(A)**), inter-contrast motion disrupts the spatial correspondence between the learned anatomical representation and the target contrast. MAX-Initial captures the target contrast appearance but fails to recover fine anatomy; consequently, the refinement receives little structural guidance and behaves largely as a single-contrast unrolled reconstruction, leaving residual artifacts and blurred boundaries. MAX with MoCo (second row, **Figure 6(A)**) restores reference-target correspondence, allowing MAX-Initial to provide a reliable starting point for refinement and yielding clearer boundaries with fewer artifacts. **Figure 6(B)** shows markedly improved reference-target alignment with MoCo. The estimated rigid motion includes translations of approximately $(-3.7, -3.0, -0.3)$ mm and rotations of $(0.3°, 0.4°, -29.4°)$, confirming the effectiveness of MoCo under large inter-contrast motion. **Figure S1** compares MAX with MoCo (MAX-MoCo) against single-contrast methods across five subjects with displaced MPRAGE references. MAX-MoCo yields the highest median PSNR and SSIM and the smallest inter-subject variation, indicating robust use of anatomical information despite inter-contrast motion.

**Figure 7** evaluates the robustness of multi-contrast reconstruction methods to reference-target structural heterogeneity. A simulated lesion (yellow arrows, **Figure 7(A)**) is present only in the target contrast and absent from the reference. dTVMRI and CDLMRI partially recover the lesion but exhibit noticeable reconstruction errors, whereas MAX-Initial fails to depict the lesion due to the absence of this target-specific abnormality from the learned anatomical representation. The subsequent physics-guided refinement successfully recovers the missing structure while preserving overall image quality, producing the smallest errors among all methods. The lesion intensity profiles along the white line (**Figure 7(B)**) show that the final MAX reconstruction most closely matches the ground truth, further demonstrating its robustness to such heterogeneity.

**Figure 8** illustrates the reconstruction results at R = 8 under different simulated SNR levels in the target measurements. As SNR decreases from 30 dB to 10 dB, the fully sampled images exhibit progressively increased noise. In contrast, MAX consistently achieves effective noise suppression while preserving anatomical structures. Even at 10 dB, MAX noticeably improves image quality, demonstrating its robustness across a wide range of SNR levels by effectively exploiting the high-SNR anatomical representations from the reference contrast.

### 4.3 Results of Ablation Studies

**Figure 9** evaluates the proposed data augmentation strategy. In **Figure 9(A)**, MAX-Single, trained on a single reference image, yields sparse embedding coverage, such that randomly sampled embeddings produce limited contrast variation. In contrast, MAX-Augmented is trained on the augmented dataset to learn a denser and more continuous embedding distribution, allowing randomly sampled embeddings to be decoded into images with diverse contrast appearances while preserving the underlying anatomy. **Figure 9(B)** shows the corresponding benefit for target-contrast adaptation: compared with MAX-Single, MAX-Augmented more accurately reconstructs the target contrast directly from the learned representation and provides a better starting point for subsequent refinement. Together, these results show that manifold expansion improves the generalization of the subject-specific representation to unseen target contrasts.

**Figure 10** compares three starting points for unrolled refinement of highly accelerated FLAIR MRI (R = 12). The zero-filled image contains severe aliasing, providing a poor starting point and leaving residual artifacts after refinement. Direct use of the fully sampled reference provides an aliasing-free but contrast-mismatched starting point. Consequently, the refined reconstruction shows suppressed artifacts but fails to accurately recover fine target-specific details. In contrast, MAX leverages the learned anatomical representation to provide a target-aware initialization, facilitating refinement and yielding the closest agreement with the fully sampled image. These results highlight the importance of learning disentangled anatomical representations for reconstruction across MRI contrasts.

## 5. Discussion

In this study, we proposed MAX, a subject-specific multi-contrast MRI reconstruction framework integrating manifold expansion, disentangled representation learning, and physics-guided reconstruction. Anatomy-preserving intensity augmentations expand the local contrast manifold, enabling a disentangled INR to separate shared anatomy from contrast-specific appearance. The learned anatomical representation is then incorporated into undersampled target reconstruction through INR adaptation and physics-guided refinement. Experiments on brain and knee MRI show that MAX consistently outperforms single- and multi-contrast methods, more accurately recovering subtle anatomical and pathological structures. The robustness of MAX to inter-contrast motion, structural heterogeneity, and measurement noise further supports the effectiveness of subject-specific anatomical representation learning for accelerated multi-contrast MRI.

### 5.1 Augmentation-Based Anatomical Representation Learning

Data augmentation has been widely adopted in deep learning-based computer vision to enhance data diversity, improve model generalization, and facilitate representation learning.[41] In this study, we extend this concept to subject-specific multi-contrast MRI to expand the local contrast manifold around a single reference image. Unlike conventional augmentation strategies, which primarily introduce generic perturbations, the proposed strategy is tailored to the reconstructed subject and aims to facilitate the learning of subject-specific anatomical representations that generalize across MRI contrasts.

The superior results on the brain dataset (**Figures 2, 3, 5(A), and 5(B)**) suggest that, for relatively homogeneous anatomy, low-cost intensity transformations can effectively enrich the local contrast manifold, facilitate disentangled representation learning, and improve target-contrast reconstruction at high acceleration factors. By comparison, the gains are smaller for knee MRI (**Figures 4 and 5(C)**). Moreover, the errors of MAX-Initial in **Figures 6 and 7** are concentrated primarily around the skull region. These observations indicate that the proposed intensity-based augmentations capture only part of the cross-contrast relationship. More complex and spatially heterogeneous contrast mechanisms may therefore require more expressive augmentation strategies. Future work could explore physics-informed contrast simulation based on Bloch equations and sequence signal models,[42–45] diffusion-model-based synthesis,[46,47] and foundation-model-based prior guidance[48] to improve subject-specific anatomical representation learning.

### 5.2 Complementary Roles of the Learned Anatomical Representation and Physics-Guided Refinement

A key challenge when using a reference image is to exploit its anatomical information without suppressing target-specific features. In MAX, INR adaptation to the target measurements and subsequent unrolled refinement play complementary roles in achieving this balance. Specifically, the adapted INR provides a target-contrast estimate constrained by the learned anatomical representation and largely free of aliasing artifacts. Building on this estimate, the unrolled refinement uses a learned CNN to recover target-specific information supported by the acquired measurements. **Figure 7** shows that a simulated lesion absent from the reference image and MAX-Initial can be recovered after refinement while preserving the surrounding tissue structures. Conversely, **Figure 8** shows that, when the target measurements are corrupted by noise, increasing the INR-anchor weight keeps refinement closer to the INR estimate and limits noise propagation. These findings show that INR adaptation and unrolled refinement complement each other by preserving reliable reference-derived anatomy while selectively incorporating target-specific information according to measurement reliability.

However, the current coupling between the INR and unrolled refinement remains indirect. The INR influences refinement through the image-domain initial estimate and a globally weighted INR-anchored regularization term; the unrolled blocks do not directly incorporate INR-derived anatomical features or account for their spatially varying reliability. Consequently, subsequent CNN and data-consistency updates may gradually deviate from reliable INR information, potentially diminishing the benefit of the learned representation and reintroducing artifacts through scan-specific overfitting. Future work could establish tighter feature-level integration by injecting INR-derived features into each unrolled block. Spatial confidence maps for INR adaptation could also be estimated to modulate the CNN updates and anchor strength. Such reliability-aware integration could preserve high-confidence anatomy, enable data-supported recovery of structures absent from the reference while suppressing unreliable updates.

### 5.3 Anatomical Representation Learning for MRI Inverse Problems

Although MAX is demonstrated for accelerated MRI reconstruction in this study, the proposed framework provides a general strategy for solving MRI inverse problems given a high-quality reference scan. Rather than directly transferring image intensities or designing handcrafted priors from the reference image, MAX explicitly learns a subject-specific anatomical representation that can be incorporated into different physics-guided reconstruction frameworks by adapting the corresponding forward model and data-consistency term. This work demonstrates its applicability to fast MRI and provides preliminary evidence of its potential for denoising (**Figure 8**). More broadly, the same framework could be extended to other MRI inverse problems, including fast quantitative MRI,[49–54] super-resolution,[55,56] and distortion correction.[57,58] By combining subject-specific anatomical representation learning with physics-guided optimization, MAX may further benefit a broad range of MRI applications, such as diffusion MRI, fMRI, longitudinal imaging, and multinuclear MRI.

### 5.4 Limitations

This study has several limitations. First, MAX introduces additional computational cost from anatomical representation learning and iterative optimization, resulting in longer reconstruction times. Optimized implementations and faster iterative solvers will therefore be important for improving computational efficiency. Second, the performance of MAX may depend on the quality of the reference image. Although this study investigated inter-contrast motion and anatomical mismatch, the effects of other degradations, including aliasing artifacts, low resolution, and geometric distortion, remain to be evaluated. Future work should also explore strategies for selectively extracting reliable information from degraded reference data while preventing corrupted components from propagating into the target reconstruction.

## 6. Conclusion

We proposed a subject-specific multi-contrast MRI reconstruction framework that learns anatomical representations from a reference image and incorporates them into physics-guided reconstruction. The proposed method improved reconstruction quality in brain and knee MRI, particularly at high acceleration rates, while maintaining robustness to inter-contrast motion, reference-target heterogeneity, and noise. These results suggest that manifold expansion-based disentangled representation learning provides an effective strategy for leveraging high-quality reference scans in MRI.

## 7. ACKNOWLEDGEMENTS

The research reported in this publication was supported by the National Institute of Biomedical Imaging and Bioengineering under Award Number R21EB031185, the National Institute of Arthritis and Musculoskeletal and Skin Diseases under Award Numbers R01AR081344, R01AR079442, and R56AR081017.

## Figure Captions

**Figure 1.** Schematic illustration of the proposed MAX framework. (A) Conceptual overview. MAX models multi-contrast MRI images as a local manifold parameterized by shared anatomical and contrast-specific representations. Anatomy-preserving data augmentation expands the local manifold from a single fully sampled reference image, enabling disentangled representation learning. The learned anatomical representation is then incorporated into a physics-guided framework for reconstructing accelerated target contrast. (B) Detailed architecture of MAX. Stage 1: A disentangled implicit neural representation (INR) is trained on augmented images generated from the fully sampled reference. Spatial coordinates are encoded into anatomical embeddings, while contrast coordinates are encoded into contrast embeddings. The embeddings are concatenated and decoded by an MLP to reconstruct the augmented images, learning shared anatomy while separating contrast-dependent appearance. Stage 2: The anatomical encoder is fixed, whereas the contrast encoder, magnitude decoder, and phase branch are adapted to the undersampled target k-space to produce an initial reconstruction, which is further refined by an unrolled physics-guided network.

**Figure 2.** Representative T2-weighted reconstruction results for two brain subjects at an acceleration factor of R = 8, using a fully sampled MPRAGE image as the reference contrast. Enlarged views of the regions outlined by the red boxes and the corresponding error maps are shown for detailed comparison. MAX provides reconstructions that are visually closest to the fully sampled images while preserving fine anatomical structures and reducing reconstruction errors.

**Figure 3.** Representative FLAIR reconstruction results for two brain subjects at an acceleration factor of R = 8, using a fully sampled MPRAGE image as the reference contrast. Enlarged views of the regions outlined by the red boxes and the corresponding error maps are shown for detailed comparison. MAX best preserves sharp tissue boundaries in both the deep gray matter and cortical regions while exhibiting the lowest reconstruction errors.

**Figure 4.** Representative knee MRI reconstruction results for two subjects with osteoarthritis at an acceleration factor of R = 6, using a fully sampled non-fat-suppressed T1-weighted image as the reference contrast. Enlarged views of the regions outlined by the red boxes and the corresponding error maps are shown for detailed comparison. In Example 1, the yellow arrows indicate subchondral cysts of the tibia, the orange arrows indicate superficial cartilage fissuring, and the blue arrows highlight the sharp surrounding anatomical boundaries preserved by MAX. In Example 2, the green arrows indicate a macerated and extruded medial meniscus. Compared with the competing methods, MAX most faithfully preserves the pathological structures while suppressing aliasing artifacts and exhibiting the lowest reconstruction errors.

**Figure 5.** Statistical comparison of reconstruction performance across all subjects. (A-C) Violin plots summarize the PSNR and SSIM distributions for brain T2-weighted, brain FLAIR, and knee MRI reconstruction, respectively. Each circle represents an individual subject, the black dot indicates the median, and the box denotes the interquartile range. MAX yields the highest median PSNR across all three tasks and the highest median SSIM for both brain contrasts. (D) Radar plot summarizing the reader study on the knee dataset across four evaluation criteria. MAX receives consistently high reader ratings across all four evaluation criteria.

**Figure 6.** Evaluation of the proposed motion-correction (MoCo) strategy under inter-contrast rigid motion. MAX-Initial denotes the INR reconstruction obtained before the unrolled refinement. (A) Reconstruction results with and without MoCo using a motion-displaced MPRAGE image as the reference contrast. Enlarged views of the regions outlined by the red boxes are shown for detailed comparison. Without MoCo, MAX-Initial fails to accurately leverage the learned anatomical representation because of inter-contrast misalignment, whereas the proposed MoCo restores anatomical correspondence, enabling accurate initialization and improved final reconstruction. (B) Spatial alignment between the reference and target contrasts with and without motion correction, together with the estimated rigid transformation between the original and motion-displaced coordinate frames.

**Figure 7.** Evaluation of the proposed framework under inter-contrast structural heterogeneity. MAX-Initial denotes the INR reconstruction obtained before the unrolled refinement. (A) Reconstruction results when a simulated lesion is introduced only in the target contrast and is absent from the reference image. Enlarged views of the regions outlined by the red boxes and the corresponding error maps are shown for detailed comparison. The yellow arrows indicate the simulated lesion. Because this target-specific structure is absent from the learned anatomical representation, MAX-Initial does not recover it. The subsequent physics-guided refinement successfully restores the lesion with clear boundary delineation. (B) Intensity profiles along the white line in (A), demonstrating that MAX most closely matches the fully sampled target and accurately recovers the simulated lesion.

**Figure 8.** Evaluation of the proposed framework under different target SNR levels. Representative

reconstruction results are shown for T2-weighted (top) and FLAIR (bottom) images at an acceleration factor of R = 8, using a high-SNR MPRAGE image as the reference contrast. Complex Gaussian noise was first added to the target k-space to simulate SNR levels of 30, 20, and 10 dB, followed by retrospective undersampling using the sampling mask shown on the left. Despite progressively increasing noise contamination in the target measurements, MAX consistently produces high-quality reconstructions with effective noise suppression while preserving fine anatomical structures.

**Figure 9.** Evaluation of the effectiveness of the proposed data augmentation strategy. MAX-Single and MAX-Augmented denote two variants of Stage 1: trained on a single reference image and the augmented image set, respectively. (A) Learned contrast embeddings and images synthesized from randomly sampled contrast embeddings with the same anatomical representation. Compared with MAX-Single, MAX-Augmented learns a richer embedding distribution, enabling more diverse contrast synthesis while preserving the underlying anatomy. (B) Target-contrast INR reconstruction results using the representations learned by MAX-Single and MAX-Augmented. MAX-Augmented provides more accurate target-contrast adaptation, resulting in improved reconstruction quality and higher PSNR/SSIM.

**Figure 10.** Evaluation of different initial estimates for the unrolled physics-guided refinement under an extreme acceleration factor of R = 12. The initial estimates obtained using zero-filled image, reference-contrast image, and the proposed MAX initialization are shown together with the corresponding final reconstructions after refinement and the associated error maps. Compared with zero-filled and reference-image initialization, MAX initialization provides a substantially more accurate initial estimate, leading to improved final reconstruction with reduced reconstruction errors relative to the fully sampled image.

**Data S1.** Local Linear Interpretation of the Learned Manifold Parameterization

**Data S2.** Error Analysis of Physics-Guided Unrolled Refinement.

**Table S1.** Summary of the base intensity transformations used to generate augmented images for manifold expansion. $\mathbf{x}$ denotes the input image, and $\mathbf{x}_\alpha$ denotes the transformed image after applying one intensity transformation. $\mathcal{U}(a,b)$ denotes a continuous uniform distribution over the interval [a, b], from which the corresponding transformation parameter is randomly sampled.

**Table S2.** Summary of quantitative reconstruction performance on each dataset. Values are reported as mean$\pm$standard deviation across test cases. The best result for each dataset and metric

is highlighted in bold.

**Figure S1.** Statistical comparison of motion-corrected MAX (MAX-MoCo) with L1-ESPIRiT, IMJENSE, and ZS-SSDU on the brain FLAIR dataset. Violin plots show the distribution of PSNR and SSIM across five subjects, with embedded box plots indicating the median and interquartile range and overlaid points showing individual cases. MAX-MoCo achieves consistently higher PSNR and SSIM than the compared methods.

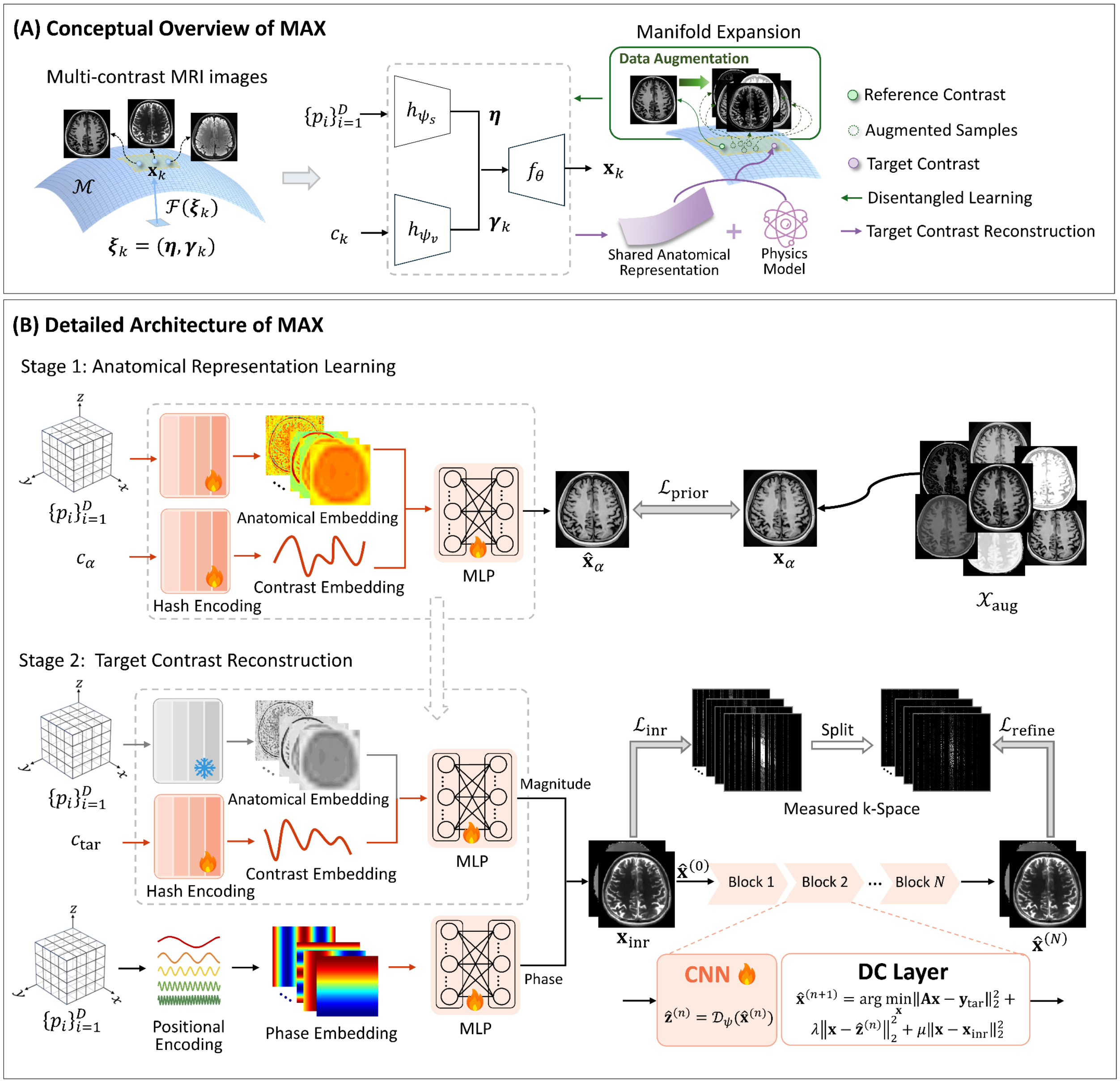

**Figure 1.** Schematic illustration of the proposed MAX framework. (A) Conceptual overview. MAX models multi-contrast MRI images as a local manifold parameterized by shared anatomical and contrast-specific representations. Anatomy-preserving data augmentation expands the local manifold from a single fully sampled reference image, enabling disentangled representation learning. The learned anatomical representation is then incorporated into a physics-guided framework for reconstructing accelerated target contrast. (B) Detailed architecture of MAX. Stage 1: A disentangled implicit neural representation (INR) is trained on augmented images generated from the fully sampled reference. Spatial coordinates are encoded into anatomical embeddings, while contrast coordinates are encoded into contrast embeddings. The embeddings are concatenated and decoded by an MLP to reconstruct the augmented images, learning shared anatomy while separating contrast-dependent appearance. Stage 2: The anatomical encoder is fixed, whereas the contrast encoder, magnitude decoder, and phase branch are adapted to the undersampled target k-space to produce an initial reconstruction, which is further refined by an unrolled physics-guided network.

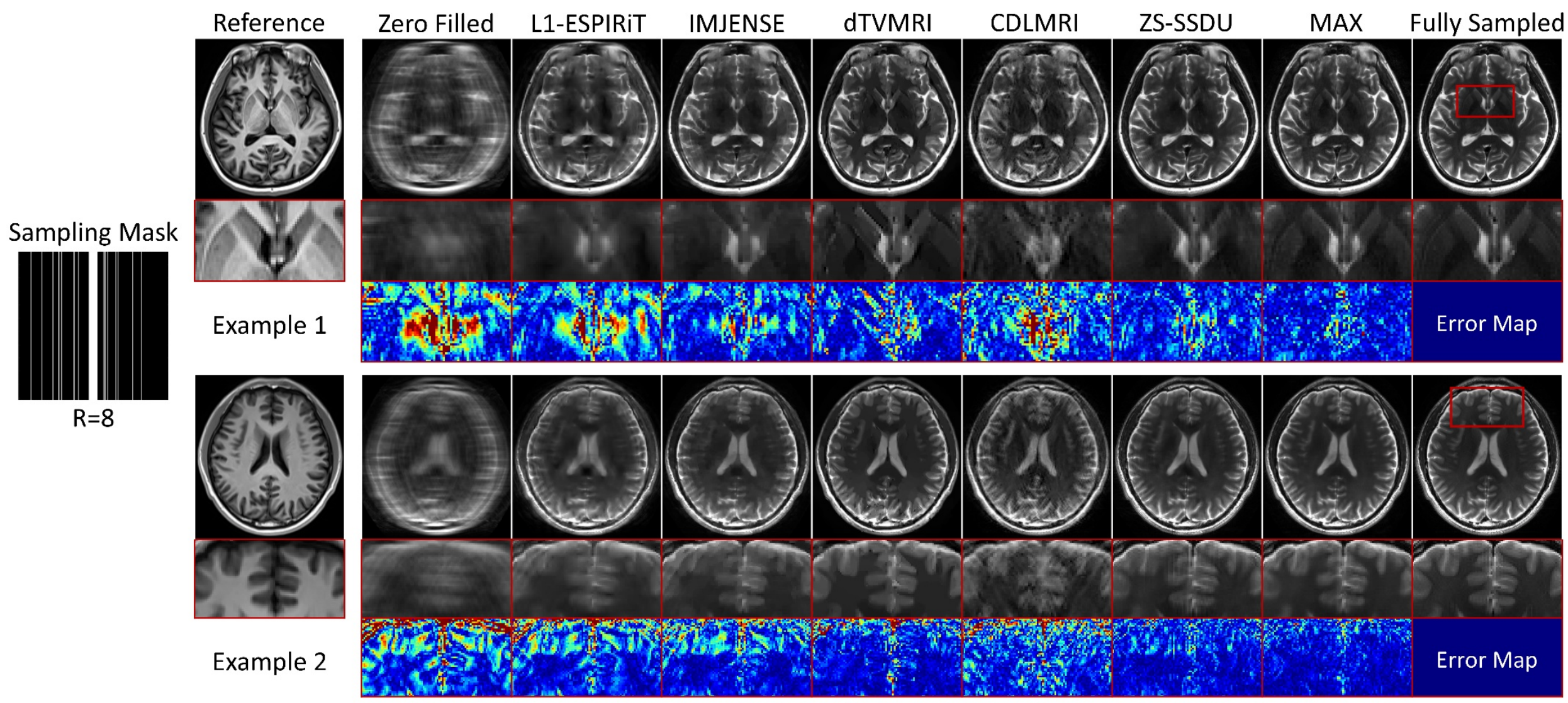

**Figure 2.** Representative T2-weighted reconstruction results for two brain subjects at an acceleration factor of R = 8, using a fully sampled MPRAGE image as the reference contrast. Enlarged views of the regions outlined by the red boxes and the corresponding error maps are shown for detailed comparison. MAX provides reconstructions that are visually closest to the fully sampled images while preserving fine anatomical structures and reducing reconstruction errors.

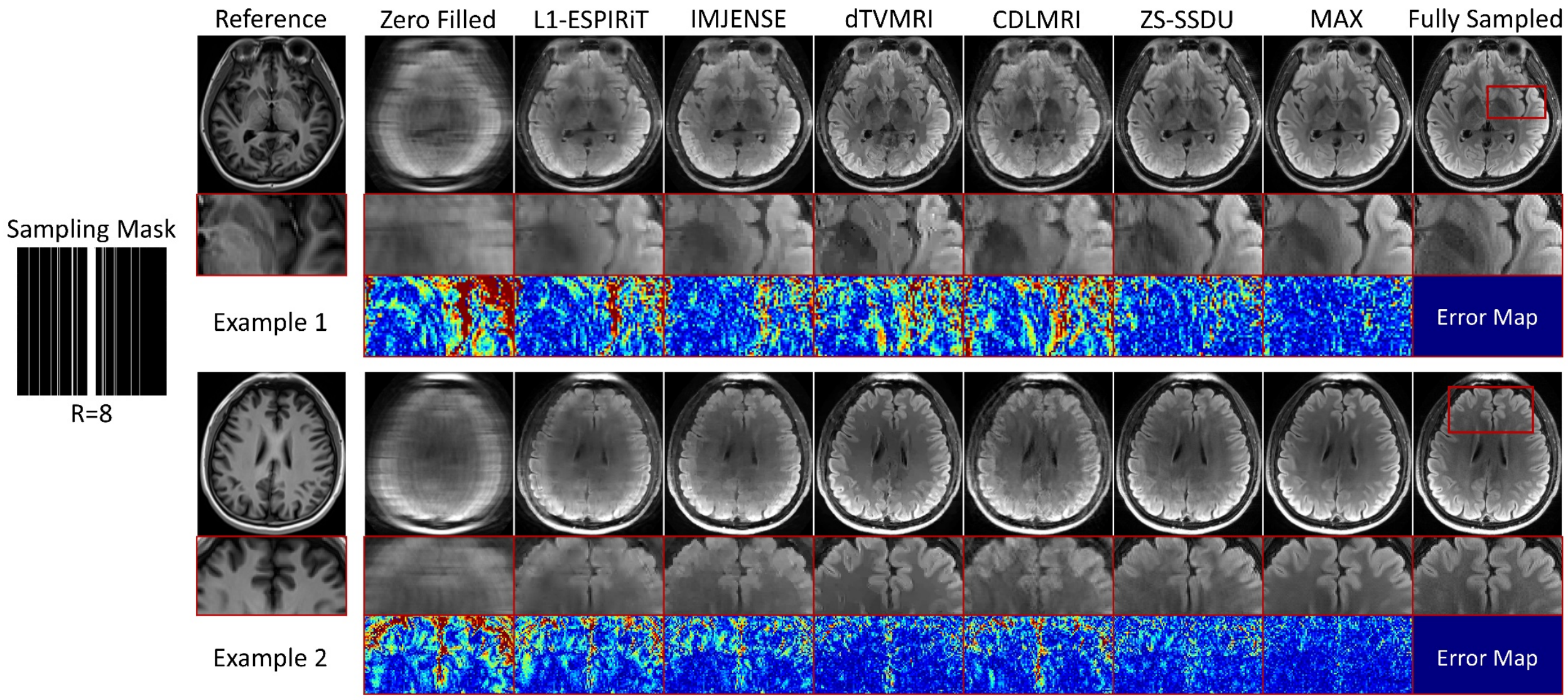

**Figure 3.** Representative FLAIR reconstruction results for two brain subjects at an acceleration factor of R = 8, using a fully sampled MPRAGE image as the reference contrast. Enlarged views of the regions outlined by the red boxes and the corresponding error maps are shown for detailed comparison. MAX best preserves sharp tissue boundaries in both the deep gray matter and cortical regions while exhibiting the lowest reconstruction errors.

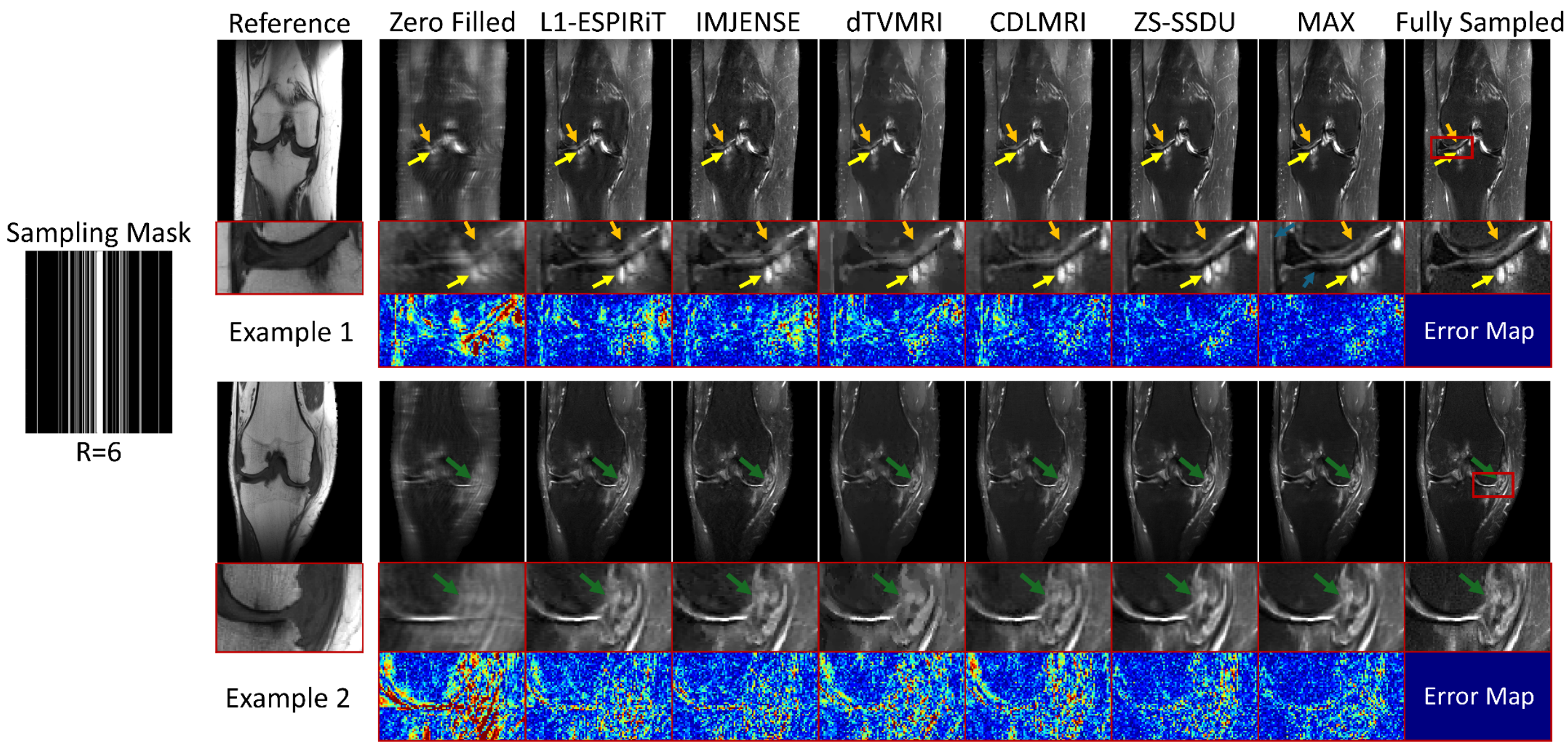

**Figure 4.** Representative knee MRI reconstruction results for two subjects with osteoarthritis at an acceleration factor of R = 6, using a fully sampled non-fat-suppressed T1-weighted image as the reference contrast. Enlarged views of the regions outlined by the red boxes and the corresponding error maps are shown for detailed comparison. In Example 1, the yellow arrows indicate subchondral cysts of the tibia, the orange arrows indicate superficial cartilage fissuring, and the blue arrows highlight the sharp surrounding anatomical boundaries preserved by MAX. In Example 2, the green arrows indicate a macerated and extruded medial meniscus. Compared with the competing methods, MAX most faithfully preserves the pathological structures while suppressing aliasing artifacts and exhibiting the lowest reconstruction errors.

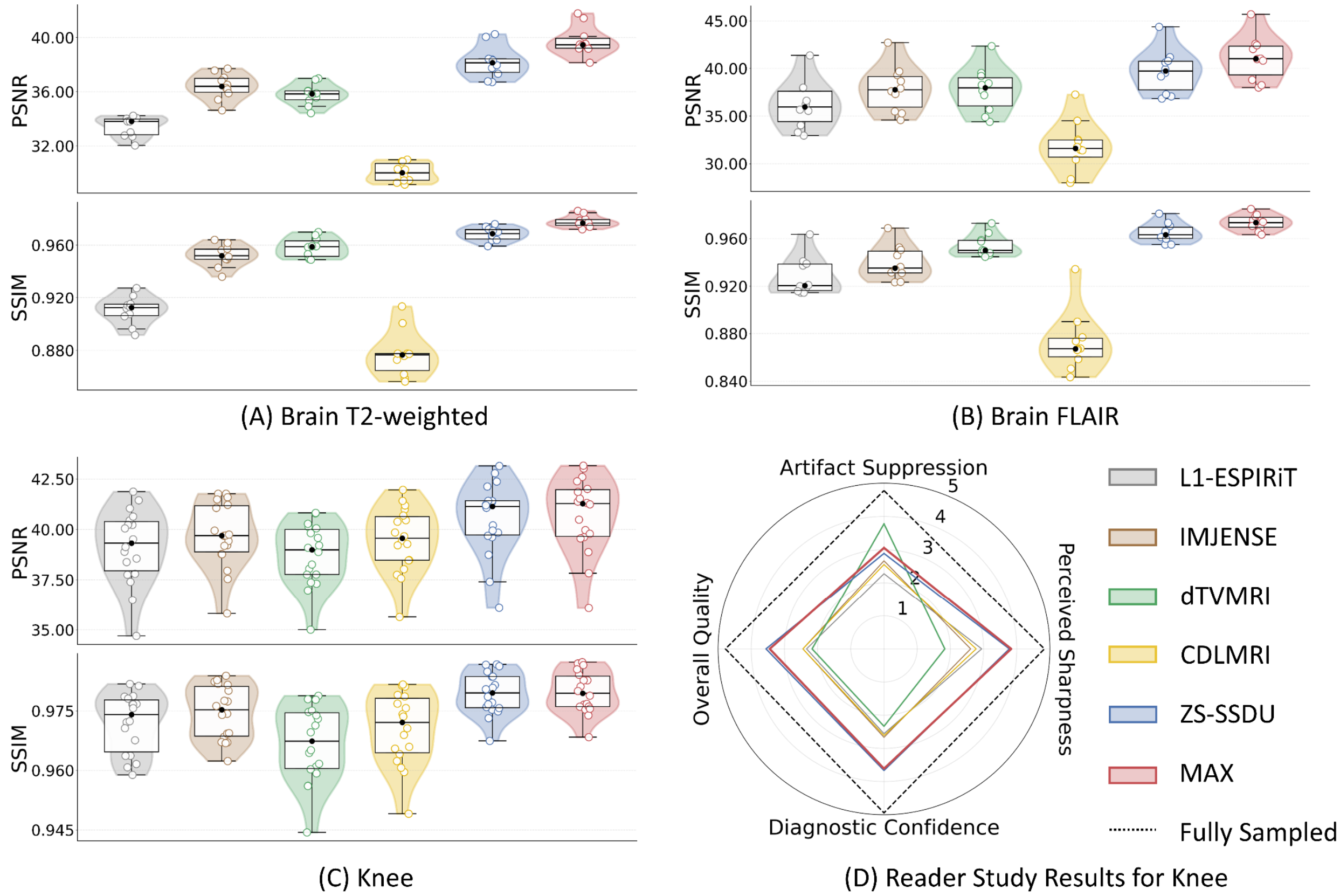

**Figure 5.** Statistical comparison of reconstruction performance across all subjects. (A-C) Violin plots summarize the PSNR and SSIM distributions for brain T2-weighted, brain FLAIR, and knee MRI reconstruction, respectively. Each circle represents an individual subject, the black dot indicates the median, and the box denotes the interquartile range. MAX yields the highest median PSNR across all three tasks and the highest median SSIM for both brain contrasts. (D) Radar plot summarizing the reader study on the knee dataset across four evaluation criteria. MAX receives consistently high reader ratings across all four evaluation criteria.

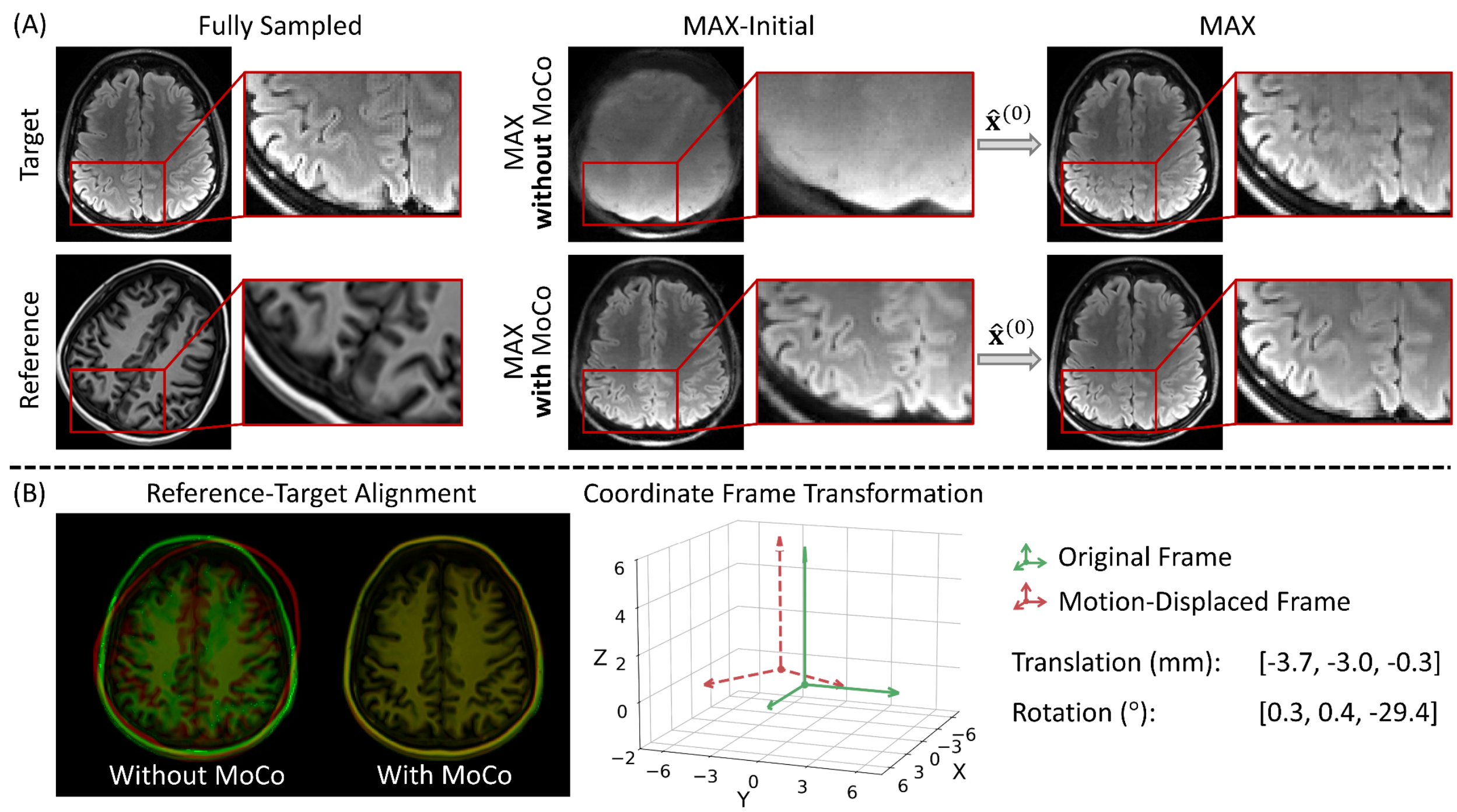

**Figure 6.** Evaluation of the proposed motion-correction (MoCo) strategy under inter-contrast rigid motion. MAX-Initial denotes the INR reconstruction obtained before the unrolled refinement. (A) Reconstruction results with and without MoCo using a motion-displaced MPRAGE image as the reference contrast. Enlarged views of the regions outlined by the red boxes are shown for detailed comparison. Without MoCo, MAX-Initial fails to accurately leverage the learned anatomical representation because of inter-contrast misalignment, whereas the proposed MoCo restores anatomical correspondence, enabling accurate initialization and improved final reconstruction. (B) Spatial alignment between the reference and target contrasts with and without motion correction, together with the estimated rigid transformation between the original and motion-displaced coordinate frames.

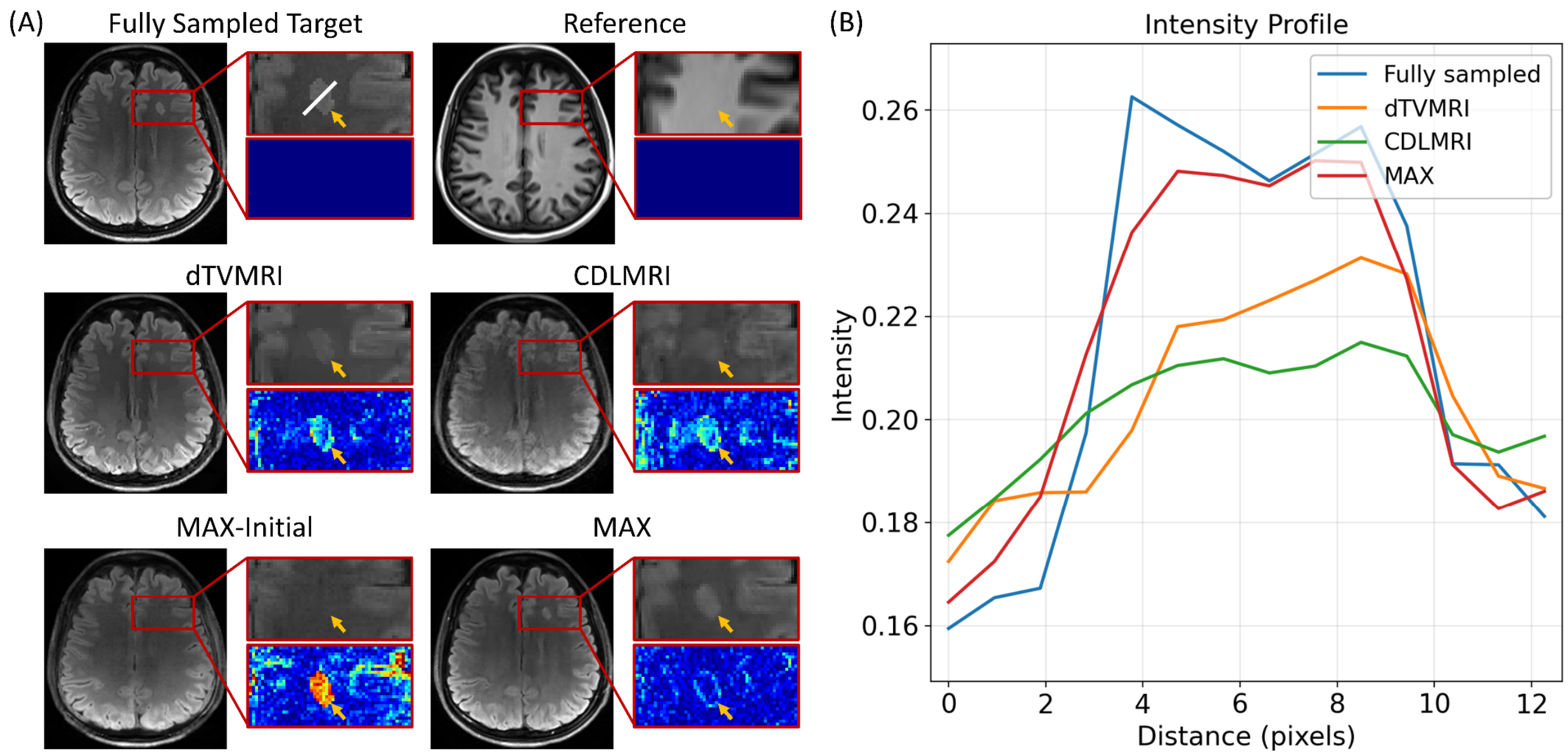

**Figure 7.** Evaluation of the proposed framework under inter-contrast structural heterogeneity. MAX-Initial denotes the INR reconstruction obtained before the unrolled refinement. (A) Reconstruction results when a simulated lesion is introduced only in the target contrast and is absent from the reference image. Enlarged views of the regions outlined by the red boxes and the corresponding error maps are shown for detailed comparison. The yellow arrows indicate the simulated lesion. Because this target-specific structure is absent from the learned anatomical representation, MAX-Initial does not recover it. The subsequent physics-guided refinement successfully restores the lesion with clear boundary delineation. (B) Intensity profiles along the white line in (A), demonstrating that MAX most closely matches the fully sampled target and accurately recovers the simulated lesion.

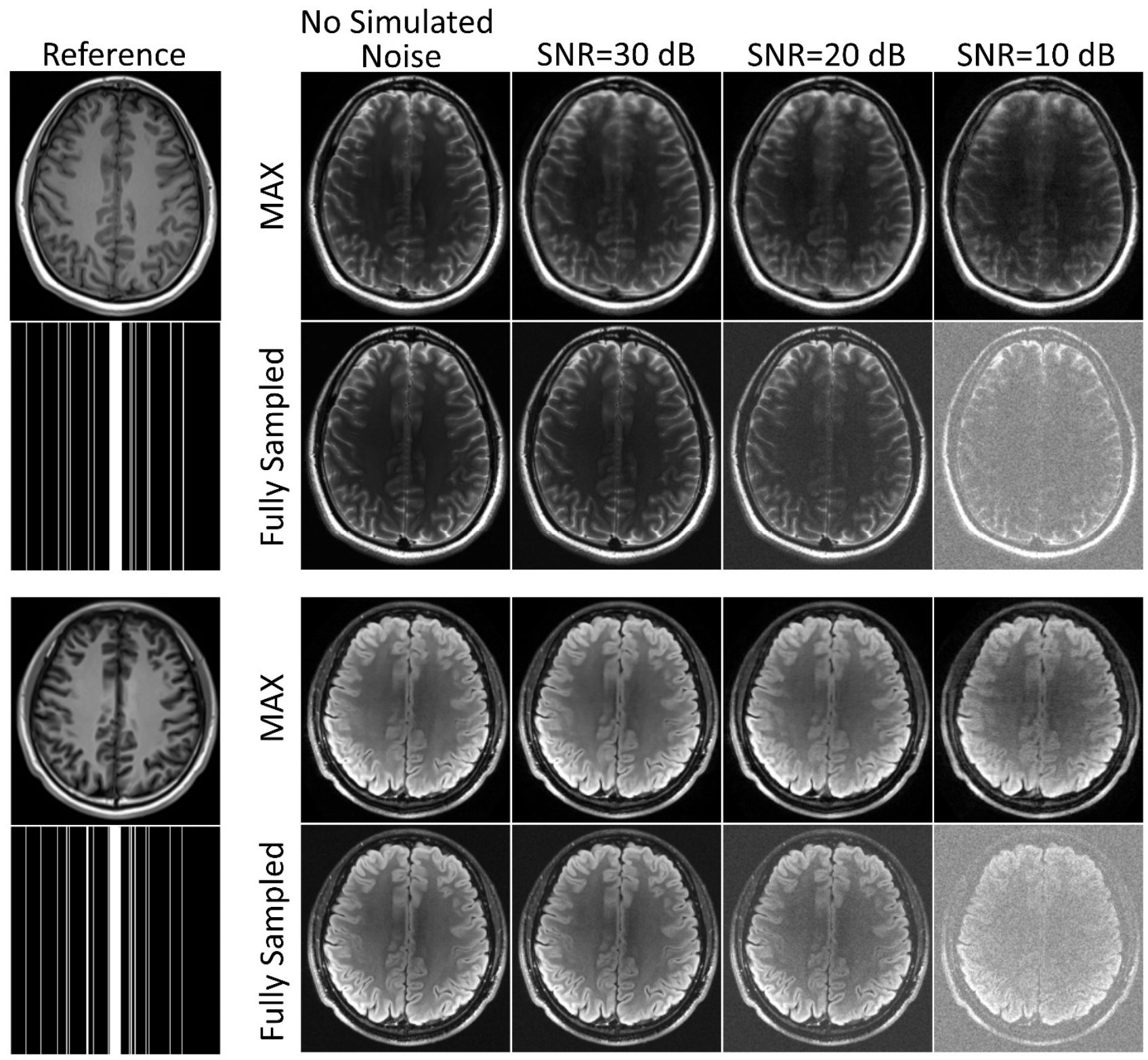

**Figure 8.** Evaluation of the proposed framework under different target SNR levels. Representative reconstruction results are shown for T2-weighted (top) and FLAIR (bottom) images at an acceleration factor of R = 8, using a high-SNR MPRAGE image as the reference contrast. Complex Gaussian noise was first added to the target k-space to simulate SNR levels of 30, 20, and 10 dB, followed by retrospective undersampling using the sampling mask shown on the left. Despite progressively increasing noise contamination in the target measurements, MAX consistently produces high-quality reconstructions with effective noise suppression while preserving fine anatomical structures.

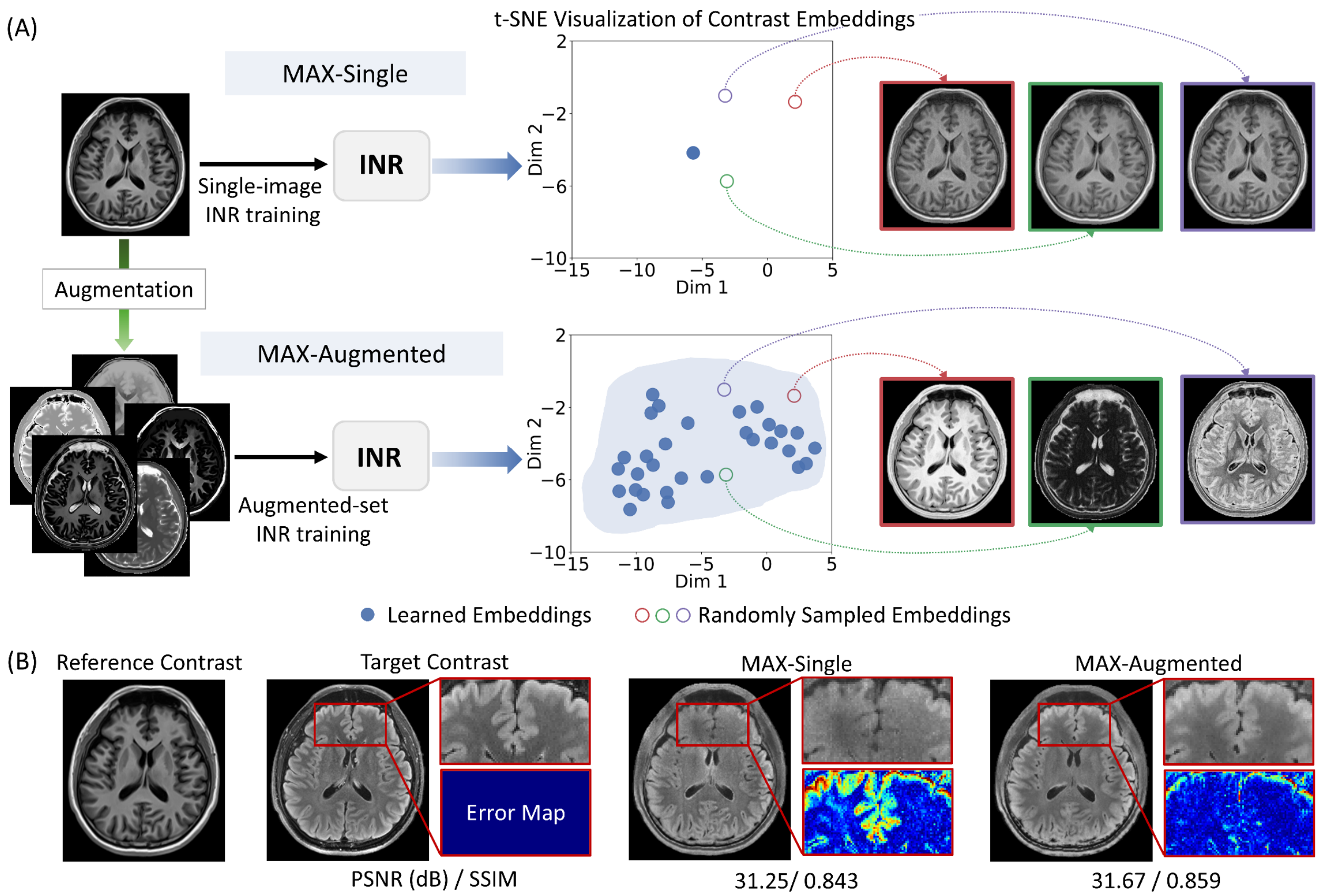

**Figure 9.** Evaluation of the effectiveness of the proposed data augmentation strategy. MAX-Single and MAX-Augmented denote two variants of Stage 1: trained on a single reference image and the augmented image set, respectively. (A) Learned contrast embeddings and images synthesized from randomly sampled contrast embeddings with the same anatomical representation. Compared with MAX-Single, MAX-Augmented learns a richer embedding distribution, enabling more diverse contrast synthesis while preserving the underlying anatomy. (B) Target-contrast INR reconstruction results using the representations learned by MAX-Single and MAX-Augmented. MAX-Augmented provides more accurate target-contrast adaptation, resulting in improved reconstruction quality and higher PSNR/SSIM.

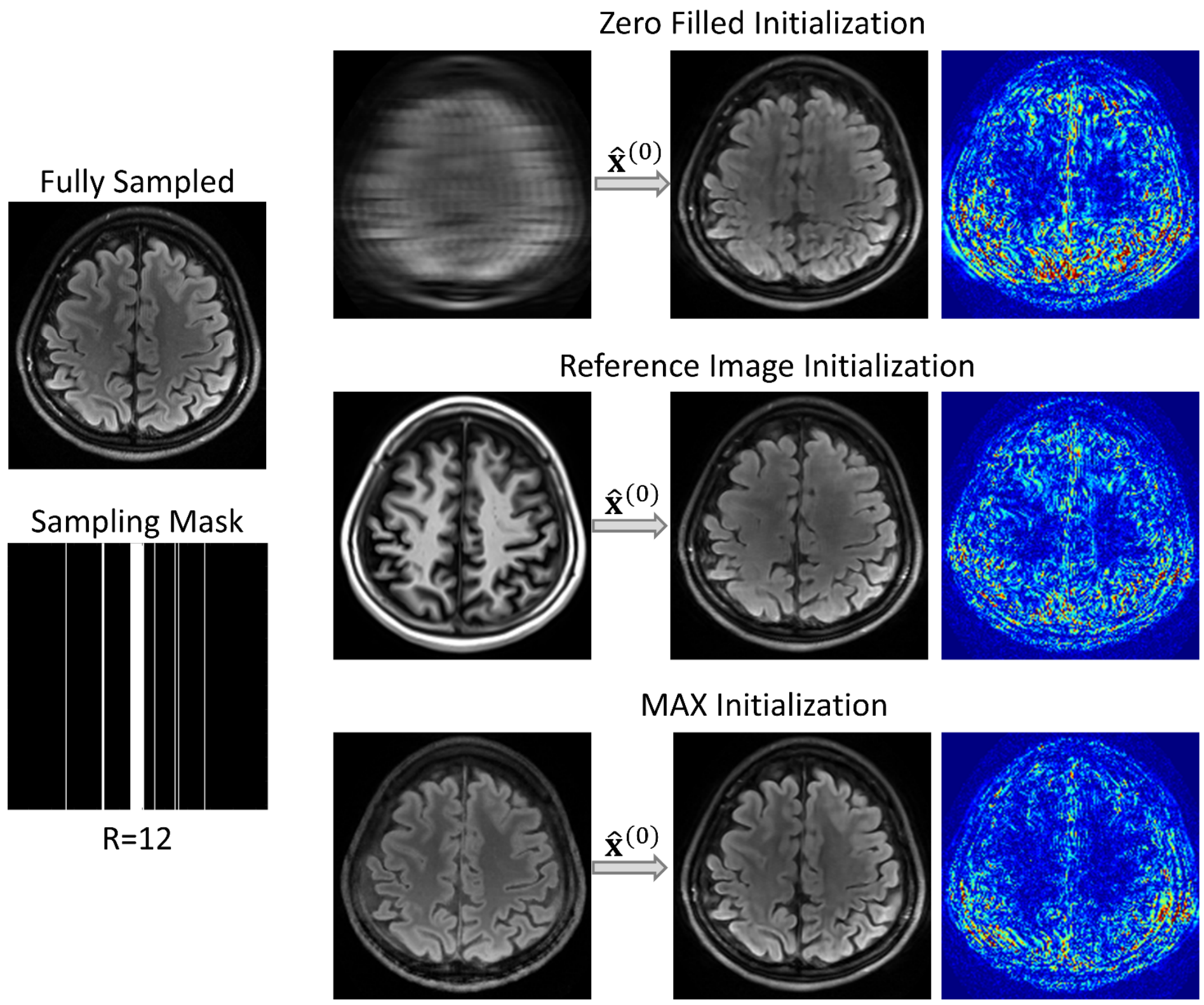

**Figure 10.** Evaluation of different initial estimates for the unrolled physics-guided refinement under an extreme acceleration factor of R = 12. The initial estimates obtained using the zero-filled image, the reference-contrast image, and the proposed MAX initialization are shown together with the corresponding final reconstructions after refinement and the associated error maps. Compared with zero-filled and reference-image initialization, MAX initialization provides a substantially more accurate initial estimate, leading to improved final reconstruction with reduced reconstruction errors relative to the fully sampled image.